\documentclass[preprint,12pt]{elsarticle}

\usepackage{amssymb}
\usepackage{amsmath}

\usepackage{amsmath,amssymb,amsfonts}
\usepackage{algorithmic}
\usepackage{graphicx}
\usepackage{textcomp}
\usepackage{xcolor}

\usepackage[T1]{fontenc}
\usepackage{booktabs}
\usepackage[misc]{ifsym}

\usepackage{dsfont}
\usepackage{amsmath,amssymb,amsfonts}
\usepackage{hyperref}

\journal{Energy Conversion and Management}

\begin{document}

\begin{frontmatter}



\title{Probabilistic Multi-Regional Solar Power Forecasting with Any-Quantile Recurrent Neural Networks\\}





\author[a0]{Slawek Smyl}

\author[a1]{Pawe\l\  Pe\l ka}
\author[a1,a2,a3]{Grzegorz Dudek}

\address[a0]{Walmart, USA, e-mail: slaweks@hotmail.co.uk}

\address[a1]{Faculty of Electrical Engineering, Czestochowa University of Technology, Al. AK 17,\linebreak   42-201 Czestochowa, Poland,  e-mail: pawel.pelka,grzegorz.dudek@pcz.pl}

\address[a2]{Faculty of Mathematics and Computer Science, University of Łódź, ul. Banacha 22, 90-238 Łódź, Poland}

\address[a3]{CAMINO - Centre for Data Analysis, Modelling and Computational Sciences, University of Łódź, ul. Narutowicza 68, 90-136 Łódź, Poland}

\begin{abstract}

The increasing penetration of photovoltaic (PV) generation introduces significant uncertainty into power system operation, necessitating forecasting approaches that extend beyond deterministic point predictions. This paper proposes an any-quantile probabilistic forecasting framework for multi-regional PV power generation based on the Any-Quantile Recurrent Neural Network (AQ-RNN). The model integrates an any-quantile forecasting paradigm with a dual-track recurrent architecture that jointly processes series-specific and cross-regional contextual information, supported by dilated recurrent cells, patch-based temporal modeling, and a dynamic ensemble mechanism.

The proposed framework enables the estimation of calibrated conditional quantiles at arbitrary probability levels within a single trained model and effectively exploits spatial dependencies to enhance robustness at the system level. The approach is evaluated using 30 years of hourly PV generation data from 259 European regions and compared against established statistical and neural probabilistic baselines. The results demonstrate consistent improvements in forecast accuracy, calibration, and prediction interval quality, underscoring the suitability of the proposed method for uncertainty-aware energy management and operational decision-making in renewable-dominated power systems.

\end{abstract}


\end{frontmatter}



\section{Introduction}

Rapid socio-economic development and the growing global demand for electricity have intensified environmental challenges related to fossil fuel dependence and greenhouse gas emissions. In response, many countries have introduced policies promoting renewable energy sources (RES), leading to a profound transformation of the energy sector. Among RES technologies, solar energy has emerged as one of the most promising and rapidly expanding options due to its abundance, sustainability, and the declining costs of photovoltaic (PV) technology. Solar power has exhibited the fastest growth rate among all renewable sources, with the total installed PV capacity increasing by 32\% in 2024 and its global share reaching approximately 6.7\% \cite{Rap25}.

However, the inherent variability and uncertainty of solar power generation pose serious challenges for power system stability, market operations, and grid integration \cite{Gan20}. A notable example is the large-scale blackout in the Iberian Peninsula in April 2025, which highlighted insufficient reserve capacity and limited grid flexibility under conditions of high renewable energy penetration. The PV output is strongly influenced by meteorological factors such as solar irradiance, temperature, cloud cover, wind speed, and aerosol concentration, which fluctuate across multiple temporal and spatial scales. As a result, PV generation is highly intermittent and stochastic, making the integration of solar energy into power grids increasingly complex. Inaccurate forecasts can lead to costly imbalances between supply and demand, inefficient reserve scheduling, and system reliability issues. Therefore, accurate and reliable forecasting of PV power output is not merely a technical necessity but a fundamental requirement for ensuring power system security, operational efficiency, and market stability.

Forecasting approaches in PV generation can be broadly classified into deterministic and probabilistic methods. Deterministic approaches produce a single point estimate of future PV output and are widely applied in grid planning, scheduling, and operational dispatch. While computationally efficient, they do not provide information on forecast uncertainty, which limits their suitability for risk-aware decision-making and the secure operation of power systems with high renewable penetration.

In contrast, probabilistic forecasting delivers predictive distributions or conditional quantiles of future PV generation, enabling a more comprehensive representation of uncertainty. Such forecasts are increasingly important for risk-aware reserve sizing, congestion management, and secure short-term scheduling, particularly in power systems with limited flexibility and reduced inertia. By providing calibrated uncertainty estimates at arbitrary probability levels, probabilistic approaches support operational decisions that extend beyond deterministic forecasting and directly contribute to improved reliability and efficiency of energy system operation.

In this study, we contribute to the advancement of PV generation forecasting by proposing a novel deep learning architecture that combines any-quantile probabilistic forecasting with a state-of-the-art recurrent neural network (RNN) specifically designed for multi-regional PV generation forecasting. The proposed model, termed the Any-Quantile Recurrent Neural Network (AQ-RNN), enables the estimation of conditional quantiles at any probability level without retraining and incorporates cross-regional contextual information to enhance forecast accuracy and calibration. Comprehensive empirical evaluations across multiple European regions demonstrate that the proposed approach consistently outperforms established probabilistic baselines, providing improved uncertainty quantification and tangible benefits for operational energy management and grid reliability.

\subsection{Related Work}

Research on PV power generation forecasting has progressed rapidly, driven by the growing demand for accurate short-term predictions to support grid operation and renewable energy integration. Early studies employed statistical and physical approaches \cite{May21,Kos19} -- such as auto-regressive models, persistence forecasting, and irradiance-based conversion chains -- but these methods often struggle with non-linear, non-stationary, and site-specific dynamics. 
Researchers continue to explore traditional approaches in search of more effective solutions; for example, \cite{Wan25a} employs sample-wise graph-based models to capture complex dependencies in PV generation data while maintaining interpretability.

\subsubsection{State-of-the-Art Machine Learning Models}

With the advent of machine learning (ML) and deep learning (DL), a large body of work has emerged focusing on data-driven PV power forecasting \cite{Luo25,Wan19,Pio24}. The vast majority of studies address deterministic forecasting. ML and DL models have gained popularity due to their strong ability to capture non-linear relationships, automatically extract hierarchical features, and adapt to large volumes of heterogeneous meteorological and power data. In contrast to conventional statistical models, they can generalize across sites and dynamically learn temporal dependencies from high-dimensional time series, enabling significant improvements in both short- and medium-term PV forecasting accuracy.

Recent advances in ML -- including attention mechanisms, hybrid architectures, and ensemble learning -- reflect a continuous evolution from purely data-driven prediction toward context-aware and interpretable forecasting frameworks capable of capturing uncertainty and enhancing operational reliability. The following studies exemplify state-of-the-art developments in this field:

\begin{itemize}
    \item \cite{Sco23} compares 64 ML models for campus-scale PV forecasting and demonstrate that random forest models achieve the highest accuracy among all tested algorithms while requiring the least amount of training data.

    \item \cite{Kim24} develops a transformer-based architecture, PVTransNet, for multi-step day-ahead PV power forecasting. The model leverages historical PV generation, weather observations, numerical weather predictions (NWP), and solar geometry data to predict hourly day-ahead power generation. A hybrid variant combines long short-term memory (LSTM) units with transformer networks to enhance weather feature representation.

    \item \cite{Tao24} proposes PTFNet (Parallel Temporal Feature Information Extraction Network), a multi-step PV forecasting model that integrates physical modeling variables with measured and NWP data through a dual-stream architecture. The model captures both temporal and inter-feature dependencies, achieving highly accurate 15-minute resolution forecasts suitable for real-world grid operations.

    \item \cite{Yan25} introduces a clustering-ensemble learning framework for day-ahead PV forecasting, combining deep time-series clustering with a weighted Warp-Euclidean distance metric to capture temporal and spatial correlations in NWP data. By integrating sequence-aware clustering with ensemble gradient boosting, the model significantly improves accuracy and stability compared to conventional baselines.

    \item \cite{Hon25} presents the Temporal and Environment-Informed Prediction (TEIP) framework, which enhances PV forecasting by jointly modeling environmental variability and temporal dependencies through a multi-spatial attention LSTM network. By dynamically capturing shading-induced fluctuations, TEIP achieves high prediction accuracy and robustness under diverse weather conditions, outperforming traditional time-series approaches.

    \item \cite{Cai25} proposes the Dynamic Stacking Ensemble Hybrid Model (DSEHM), which integrates hybrid deep neural networks (NNs), attention mechanisms, tree-based models, and dynamic model selection to improve forecasting of non-stationary PV power generation. The architecture combines advanced components such as Informer, AttnGRU, and TCN with SOM-based dimensionality reduction and GMM clustering for adaptive model stacking, substantially enhancing accuracy and robustness.

    \item \cite{Alm24} introduces the Multilevel Data Fusion and Neural Basis Expansion Analysis (MF-NBEA) framework for regional PV power forecasting. By integrating multi-source exogenous data, spatial information, and neural decomposition modules, MF-NBEA achieves compact and interpretable representations. Evaluated on real-world datasets, the framework outperforms state-of-the-art DL models, providing more accurate and robust forecasts that support efficient and sustainable energy management.
\end{itemize}

\subsubsection{Probabilistic Solar Energy Forecasting}


Probabilistic forecasting in power systems has a profound impact on operational security and decision-making across the entire energy sector. In contrast to point forecasting, probabilistic forecasting provides predictions in the form of probability density functions (PDFs), quantiles, or prediction intervals (PIs), thereby quantifying forecast uncertainty. Its practical relevance for grid operation and control is explored in~\cite{Hau19}. A comprehensive overview of probabilistic forecasting in smart grid contexts is provided in~\cite{Kha22}, while the integration of probabilistic solar forecasts into power system operations is extensively reviewed in~\cite{Li20}. A broader survey on probabilistic solar power forecasting methodologies can be found in~\cite{Mee18}. 

Probabilistic forecasting can be classified into three main categories: input scenario simulation, post-processing of point forecasts, and model-based probabilistic modeling \cite{Hon16}. The first approach generates multiple forecasts by varying input scenarios such as solar irradiance or temperature, which are then aggregated into probabilistic outputs. The second approach derives uncertainty estimates by modeling residuals or combining forecasts, emphasizing the importance of the underlying model’s accuracy. Finally, model-dependent methods directly estimate prediction intervals or full probabilistic distributions, often leveraging ensemble learning, deep networks, or hybrid techniques to capture nonlinear dependencies and uncertainties inherent in PV generation.

Probabilistic forecasts can be issued in forms of probability distributions, quantiles, or intervals, using parametric, semiparametric, or nonparametric approaches~\cite{Hon20}. 
A parametric approach assumes a certain PDF for the forecast distribution and involves a forecasting model that produces parameters of this PDF. For example, in \cite{Ram19} the model distinguishes between sunny, overcast, and partly cloudy conditions, assigning each a Gaussian distribution. A parametric probabilistic model proposed in \cite{Fan22} addresses the limitations of normal-distribution-based methods by incorporating fat-tailed distributions such as Laplace and two-sided power distributions. These distributions are integrated into a DeepAR RNN framework, which maps input features to the parameters of the conditional output distribution. A novel loss function based on the continuous ranked probability score (CRPS) is derived in closed form for these distributions, enabling efficient training and improved reliability in uncertainty quantification.

Semi-parametric approaches combine parametric forms with flexible, data-driven components, partly relaxing distributional assumptions.
The semiparametric models described in \cite{Fer23} for day-ahead hourly PV generation forecasts combine three deterministic forecasts as inputs with a flexible statistical layer that models the residual uncertainty without strictly fixed distributional assumptions. Weather-forecast variables from a NWP model and sun-position factors serve as explanatory inputs, while the statistical component estimates probabilistic output distributions based on the deterministic forecasts.

Nonparametric approaches avoid assumptions about the functional form of the predictive distribution, offering greater flexibility for modeling complex, data-driven uncertainty structures. Widely used techniques include quantile regression (QR) \cite{Mas24}, quantile regression forests (QRF), and clear-sky persistence ensembles (CSPE) \cite{Vis24}, alongside more sophisticated hybrid ML models. For example, \cite{Fen25} introduces a nonparametric probabilistic forecasting framework based on the distribution-free B-spline-iTransformer architecture (BS-iMCFormer), which integrates kernel density estimation (KDE) with B-splines to represent probability density functions (PDFs). The forecasting task is reformulated as predicting B-spline coefficient vectors using a cross-modal transformer, allowing for flexible, assumption-free uncertainty quantification.

A related method is proposed in \cite{Wan22}, where day-ahead interval forecasts are generated using a quantile regression LSTM (QRLSTM). Historical PV data are first clustered via $k$-means, and scenario-based features are synthesized using a deep convolutional GAN before being input into the QRLSTM to yield prediction intervals. Another quantile-based approach is presented in \cite{Ait25}, which proposes a hybrid Q-CNN-GRU model that combines convolutional neural networks and gated recurrent units to estimate multiple conditional quantiles. Trained with a quantile loss function and tested across diverse geographic locations, seasons, and forecast horizons, the model demonstrates superior accuracy, reliability, and calibration, particularly in multivariate settings incorporating exogenous meteorological variables.

An additional noteworthy example of hybrid ML for probabilistic forecasting is presented in \cite{Son25}, where advanced neural architectures -- specifically CNNs, bidirectional LSTM networks, and attention mechanisms -- are used to capture spatial and temporal dependencies in the data. These learned feature representations are then passed to a natural gradient boosting (NGBoost) model, which outputs full predictive distributions, effectively combining deep learning with probabilistic gradient boosting for enhanced uncertainty modeling.

\subsubsection{Multi-Site Forecasting}

Multi-site forecasting extends conventional single-plant prediction by integrating spatially distributed measurements across multiple photovoltaic installations. This paradigm exploits the coherent spatial propagation of cloud fields and irradiance patterns, enabling upstream sites to provide anticipatory information on conditions that will later affect downstream plants. Empirical studies confirm that incorporating these spatial dependencies yields measurable reductions in forecasting error and enhanced robustness under rapidly varying atmospheric conditions \cite{Ver25}. Furthermore, multi-site architectures enable simultaneous prediction for large plant portfolios, an essential capability for regional energy management, aggregation services, and the efficient operation of intelligent energy communities, while reducing computational cost compared to training isolated single-site models.

The methodological landscape has broadened substantially. Early contributions include linear spatio-temporal autoregressive (ST-AR) schemes applied at fleet scale, such as the short-term prediction of 303 PV plants in \cite{Car20}. Tensor-decomposition-based feature extraction followed by predictive clustering \cite{Cor20} has been shown to efficiently leverage shared structures across sites, while \cite{Sev21} introduced a fuzzy evolving clustering mechanism capable of handling concept drift and multivariate data streams in a spatio-temporal setting. DL approaches have further advanced multi-site PV forecasting. Hybrid CNN-LSTM architectures that embed spatial matrices of neighbouring sites \cite{Zan20}, together with bidirectional and attention-enhanced LSTM variants \cite{Bra20}, demonstrate consistent performance gains by jointly capturing localized spatial interactions and the temporal evolution of PV power.

A major methodological shift is the transition from grid-based to graph-based spatio-temporal modeling. Graph neural networks (GNNs) now constitute the dominant family of models due to their ability to naturally represent the irregular geographical deployment of PV plants and to propagate information over arbitrary network topologies. Notable examples include graph-convolutional irradiance prediction models \cite{Zha22}, spectral spatio-temporal GNNs for multi-target PV forecasting \cite{Sim22}, and GConvLSTM architectures that couple graph convolution with recurrent temporal modeling to provide simultaneous predictions across all sites \cite{Ver22}. More recent developments incorporate self-adaptive adjacency matrix construction combined with graph convolution and LSTM mechanisms \cite{Zan24}, thereby distinguishing stable and transient inter-site dependencies and enhancing spatial representation fidelity.

Probabilistic multi-site PV forecasting has also become an active research area. The convolutional graph autoencoder introduced by \cite{Kho20} estimates the full distribution of future irradiance by combining graph spectral convolutions with generative modeling. Building on this direction, \cite{Wan25b} proposes a dynamic graph-based probabilistic forecasting approach that jointly models shape and amplitude characteristics of PV power, thereby improving uncertainty quantification in ultra-short-term multi-site settings. A complementary development introduces a spatial-temporal multi-task learning framework based on a GraphGRU architecture \cite{Bai24}, which enhances probabilistic accuracy by jointly capturing inter-site spatial correlations and temporal dynamics.

\subsection{Motivation and Contributions}

Accurate forecasting of solar power generation is essential for maintaining grid stability, yet the strong variability and uncertainty of PV output make this task challenging. Despite significant progress in PV power forecasting, several research gaps remain that limit the operational applicability of advanced models in real-world energy systems: 

\begin{itemize}
    \item Earlier studies have primarily focused on deterministic forecasting, which fails to capture the uncertainty inherent in PV generation and is therefore insufficient for supporting informed decision-making in grid management.
    \item Most existing forecasting approaches focus on single-site predictions, whereas regional power systems with distributed PV resources require models capable of exploiting spatial dependencies across multiple locations.
    \item Although nonparametric probabilistic models offer greater flexibility than parametric approaches, they are typically constrained to a fixed set of quantile levels, reducing their adaptability.
\end{itemize}

To address these gaps, we introduce AQ-RNN, a forecasting framework that combines the flexibility of nonparametric quantile estimation with the representational power of deep recurrent architectures. The main contributions of this study are summarized below:

\begin{enumerate}
    \item \textbf{A novel Any-Quantile recurrent framework for probabilistic forecasting.}  
    We propose AQ-RNN, a new architecture capable of producing arbitrary quantile forecasts within a single model, offering a flexible and efficient solution for uncertainty estimation.

    \item \textbf{Dual-track recurrent architecture for multi-regional multi-target forecasting.} 
    The model employs a two-track design: a {primary} RNN responsible for series-specific forecasting, and a {context} RNN that extracts cross-regional information and dynamically adapts it to each forecasted series. Once trained, the model is able to generate forecasts for any quantile across all regions.
    \item \textbf{Innovative architectural components.}  
    AQ-RNN integrates several mechanisms to enhance learning efficiency and robustness:  
    (i) dilated recurrent cells expanding the temporal receptive field;  
    (ii) patching with internal context to capture short- and long-term dependencies;  
    (iii) dynamically assembled RNN teams that promote specialization and adaptivity;  
    (iv) overlapping quantile-level ranges that improve smoothness and consistency across the quantile spectrum.

    \item \textbf{Comprehensive empirical evaluation.}  
    Extensive experiments across 259 European regions compare AQ-RNN against statistical (ARIMA, Theta) and neural (Transformer, TFT, WaveNet, DeepAR) probabilistic baselines. AQ-RNN achieves the lowest CRPS, MARFE, and Winkler scores, demonstrating superior calibration, sharpness, and overall predictive performance.
\end{enumerate}

The remainder of this paper is organized as follows.  
Section~2 formulates the forecasting problem.  
Section~3 introduces the proposed neural architecture.  
Section~4 presents the empirical results.  
Finally, Section~5 concludes the paper.

\section{Forecasting Problem}

Let $\{Z_t\}$ denote the PV power generation time series with resolution $u$ (e.g., $u=24$ for hourly resolution, $u=96$ for 15-minute resolution).  
The objective is to forecast the $q$-th quantile of the next $h \cdot u$ time steps, based on the preceding $m \cdot u$ values, where $h$ is the forecast horizon and $m$ is the lookback window length, both expressed in days.

\begin{figure}[t]
\centering
\includegraphics[width=0.7\textwidth]{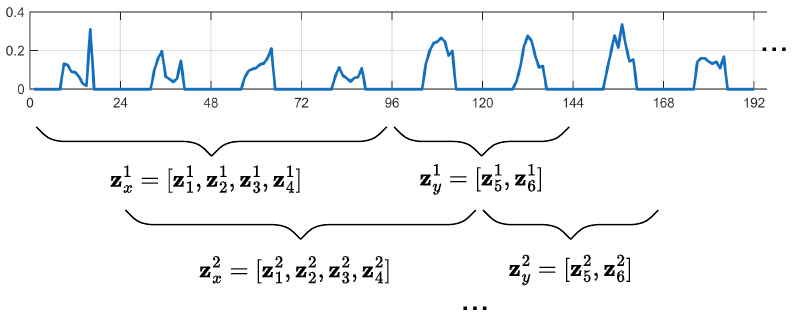}
\caption{Time-series segmentation for constructing input and output sequences.}
\label{figSeg}
\end{figure}

Define the input sequence of length $m$ days as $\mathbf{z}_x^j = [\mathbf{z}_1^j, \ldots, \mathbf{z}_m^j] \in \mathbb{R}^{m \cdot u}$ and the output sequence of length $h$ days as $\mathbf{z}_y^j = [\mathbf{z}_{m+1}^j, \ldots, \mathbf{z}_{m+h}^j] \in \mathbb{R}^{h \cdot u}$, where $\mathbf{z}_i^j = [z_{i,1}^j, \ldots, z_{i,u}^j]$ denotes the PV power values for day $i$ at time index $j = 1, 2, \ldots$.  
An example of segmenting the time series $\{Z_t\}$ into input and output sequences for $m=4$ and $h=2$ is shown in Fig.~\ref{figSeg}.

In this study, we set the forecast horizon to $h = 2$ days to align with typical operational requirements of power system management. System operators commonly prepare forecasts at least one day ahead, often with a lead time of 12 hours or more, as noted in~\cite{Pop15}.

The forecasting model $f$ produces estimates of the $q$-th quantile of $\mathbf{z}_y^j$ according to $\hat{\mathbf{y}}^j_q = f(\mathbf{z}_x^j, q; \theta)$, where $q \in (0,1)$ is the quantile level and $\theta$ denotes the model parameters.

In the multi-regional forecasting setup, the goal is to predict $q$-th quantiles of PV power output for all $L$ regions. Accordingly, the training dataset comprises historical input–output sequence pairs aggregated from each of the $L$ regions.

{This study relies exclusively on historical PV generation data and does not incorporate exogenous weather variables such as solar irradiance, temperature, or cloud cover. While numerical weather predictions are commonly used in operational settings, their availability, temporal resolution, and accuracy vary substantially across regions and over long historical periods. In particular, high-resolution hourly weather data consistently covering the entire 30-year experimental period and all 
$L = 259$ regions are not available, and practical deployment would require access to forecasted rather than observed weather variables up to 48 hours ahead. Restricting the input to generation-only data therefore enables a consistent large-scale evaluation, ensures fair comparison across forecasting models, and reflects realistic operational scenarios in which reliable future weather forecasts may be unavailable or subject to significant uncertainty.}



\section{Any-Quantile Forecasting Model} \label{sec:framework1}

\subsection{Architectural Components: Innovations and Inspirations} \label{sec:framework}

The proposed Any-Quantile Recurrent Neural Network (AQ-RNN) model is defined by the following key components and mechanisms:

\begin{enumerate}
    \item \textbf{Any-quantile probabilistic forecasting framework}. This methodology enables the model to be trained across the entire probability range $(0,1)$, allowing it to predict any quantile level during inference. Once trained, the model can produce a full conditional distribution of the output variable without the need for retraining.
    
    \item \textbf{Dual-track RNN architecture: primary and context tracks}. The primary track generates the quantile forecasts for regional solar power generation along with a confidence estimate. The context track processes auxiliary information from other regions and provides learned contextual signals that support the primary track.
    
    \item \textbf{Dilated recurrent cells}. The use of dilation extends the receptive field of the recurrent units, allowing the model to capture longer temporal dependencies without increasing the number of layers or parameters. 
       
    \item \textbf{Patching with internal context}. Temporal input sequences are divided into patches, each augmented with learned internal contextual information. These enriched patches are processed by separate RNN streams, allowing the model to specialize in capturing distinct temporal patterns within each segment.

    \item \textbf{Dynamically assembled teams}. A collection of RNN models is trained in parallel. To generate the output, the most confident team members are combined, enabling the model to adapt to varying temporal patterns.

    \item \textbf{Overlapping quantile-level ranges}. Quantile forecasts are generated over overlapping intervals of quantile levels to ensure smoothness and consistency across the full quantile spectrum. A dedicated RNN team is trained for each interval, enabling localized expertise across different distributional regions.
\end{enumerate}

Any-quantile probabilistic forecasting is a novel methodology introduced recently in \cite{Smy26}, which enables the prediction of conditional quantiles at any probability level. This approach is model-agnostic and can be applied to a wide range of ML models and forecasting tasks. In \cite{Smy26}, we implemented this framework using ESRNN (a hybrid model combining exponential smoothing and recurrent neural networks) as well as N-BEATS, to address the short-term load forecasting problem. In the present study, we apply the any-quantile methodology to the task of solar power generation forecasting using a newly developed recurrent architecture that differs fundamentally from ESRNN in both structure and internal mechanisms.

\begin{figure}[t]
\centering
\includegraphics[width=0.9\textwidth]{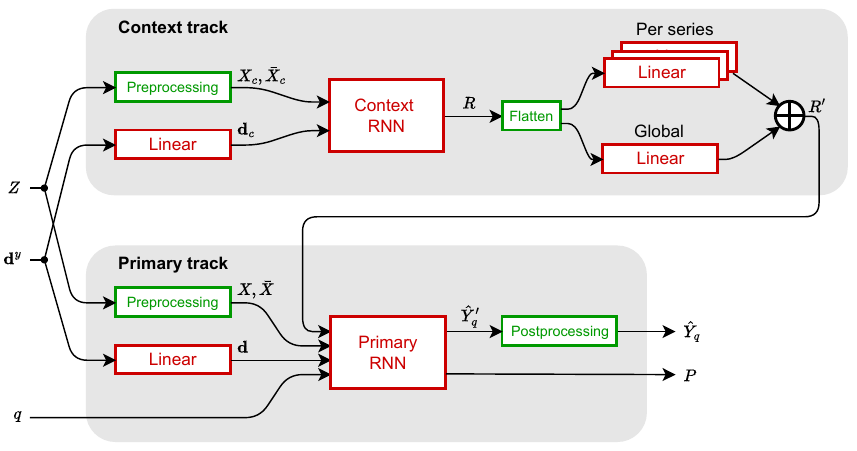}
\caption{Block diagram of the proposed AQ-RNN forecasting model.}\label{figBD}
\end{figure}

Unlike ESRNN, the proposed model features a dual-track architecture (see Fig.~\ref{figBD}). This design, based on two processing paths, primary and context, was first introduced in \cite{Smy23} and \cite{Smy24} to incorporate additional contextual information extracted dynamically, for example, from exogenous variables or from variables representing similar phenomena in other regions. In the current application, the context track processes information from all 259 regions and dynamically adapts it per series by generating a context vector, which serves as an additional input to the primary RNN, tailored to the specific series being forecasted. Unlike earlier implementations, where the context vector was adapted via a modulation mechanism, the current model uses a combination of global and per-series embeddings to adjust the context representation.

\begin{figure}[t]
\centering
\includegraphics[width=0.8\textwidth]{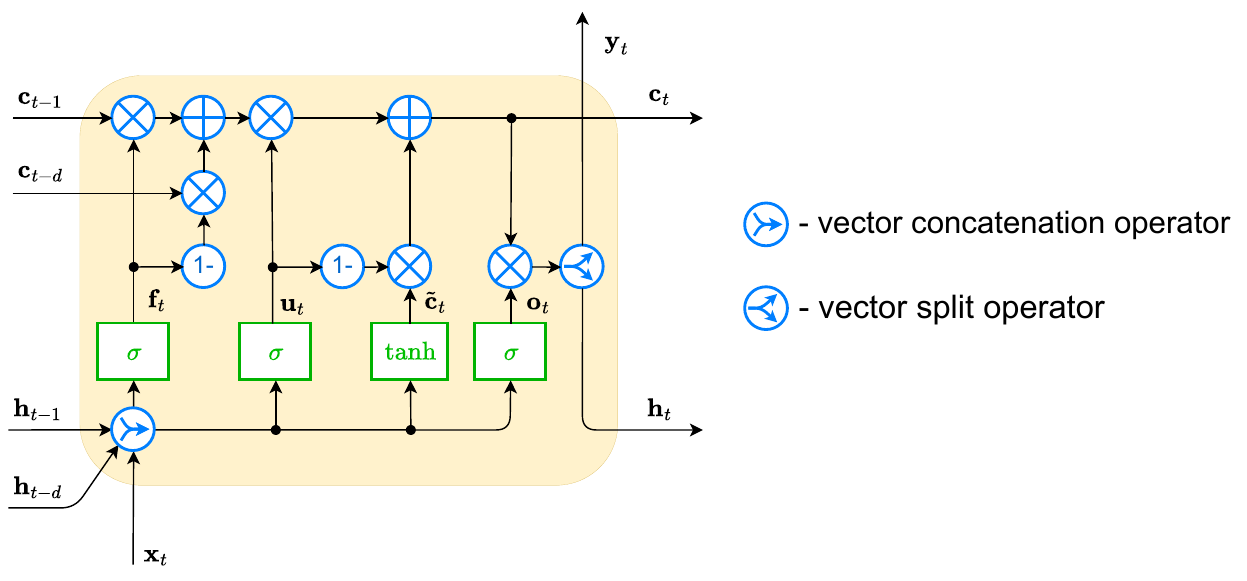}
\caption{Illustration of the dilated recurrent cell (dRNNCell).}\label{figCells}
\end{figure}

Our dual-track RNN architecture employs a novel type of recurrent cell introduced in \cite{Smy24a}: the dilated recurrent cell dRNNCell, illustrated in Fig.~\ref{figCells}. This unit serves as an alternative to traditional LSTM and GRU cells, featuring an original internal design that incorporates both recent and delayed hidden states. Such a mechanism effectively expands the model's receptive field, particularly in deep, multilayer architectures, where successive layers further extend the temporal coverage of preceding ones. This is especially advantageous for capturing seasonal and long-range dependencies in time series.

Another innovation distinguishing the dRNNCell from standard LSTM and GRU cells is the separation of its output into two streams: the real output $\mathbf{y}_t$, which is passed to the next layer and carries information relevant to the forecasted variable, and the controlling output $\mathbf{h}_t$, which serves as input to the gating mechanism at subsequent time steps.

Building on this overview, the subsequent sections present an in-depth discussion of the model's architecture and key mechanisms.

\subsection{Data Processing in the Dual-Track Architecture}

\subsubsection{Inputs and Outputs}
\label{IaO}

The AQ-RNN model is trained in a cross-learning mode \cite{Smy20}, meaning it is trained simultaneously on $L$ time series (in our case, 259 hourly time series of solar power generation). As illustrated in Fig.~\ref{figBD}, the input $Z$ represents a set of $L$ series. Both RNNs are trained on $Z$, but with different batch sizes: the context RNN processes all $L$ series in a batch, while the primary RNN processes only a subset of them. The preprocessing component segments the time series, normalizes the data, and constructs the training set. Specifically, input segments of length $m$ days and output segments of length $h$ days are defined (see Fig.~\ref{figSeg}).

The input $\mathbf{x}$ and output $\mathbf{y}$ variables are obtained by normalizing the corresponding data segments according to the following formulas:

\begin{equation}
\mathbf{x}^j = \frac{\mathbf{z}_x^j}{\bar{z}_x^j}, \qquad 
\mathbf{y}^j = \frac{\mathbf{z}_y^j}{\bar{z}_x^j},
\label{xy}
\end{equation}
where $\mathbf{x}^j = [\mathbf{x}_1^j, \ldots, \mathbf{x}_m^j]$ and $\mathbf{y}^j = [\mathbf{y}_1^j, \ldots, \mathbf{y}_h^j]$ denote the normalized input and output segments corresponding to the actual segments $\mathbf{z}_x^j = [\mathbf{z}_1^j, \ldots, \mathbf{z}_m^j]$ and $\mathbf{z}_y^j = [\mathbf{z}_{m+1}^j, \ldots, \mathbf{z}_{m+h}^j]$, respectively, with $\bar{z}_x^j$ representing the mean value of the input segment.

The mean value of the input segment, $\bar{z}_x^j$, is introduced as a separate input variable to both the context and primary RNNs, denoted as $\bar{Z}_c$ and $\bar{Z}$ in Fig.~\ref{figBD}. This allows the model to incorporate information about the local level of power generation specific to each region.

Another input variable, used to encode seasonal information, is a calendar variable $\mathbf{d}^y \in \{0,1\}^{52}$, which represents the week of the year as a one-hot encoded binary vector. This feature introduces yearly seasonality, which is important for modeling solar power generation. The binary vector is passed through a linear layer to produce a low-dimensional real-valued embedding vector $\mathbf{d}$.

In addition, the primary track receives the quantile level $q$ to be forecasted, as well as context information $R'$ provided by the context track. This context information aggregates signals extracted from all $L$ time series and assists in forecasting the individual series processed in the current batch by the primary RNN. The mechanism by which $R'$ is generated by the context track is described in a later section.

The primary RNN produces a forecast of the $q$-th quantile for the normalized data, $\mathbf{y}^{\prime j}_q \in \mathbb{R}^{h\cdot u}$, which is then denormalized in the {Postprocessing} component:

\begin{equation}
\hat{\mathbf{y}}^j_q = \hat{\mathbf{y}}^{\prime j}_q \cdot \bar{{z}}_x^j
\label{eqyq}
\end{equation}

The primary RNN also outputs a scalar variable $p$, which expresses the model's confidence level in its forecast. The role and interpretation of this confidence score are explained in subsequent sections.

\subsubsection{Extracting Context Information} \label{context}

As in our previous studies, we employ a dedicated context RNN whose role is to extract a condensed representation of the general forecasting environment. This context aims to capture information from time series that are not included in the primary RNN's input batch.

In this work, due to the relatively modest number of series ($L=259$), we are able to process the entire dataset synchronously at each time step using the context RNN. 
Specifically, the context RNN operates on a batch containing all 259 series and produces a short output vector for each of them. For example, when the per-series context length is set to 2, the resulting 259 vectors are concatenated into a single 518-dimensional context vector.

This vector is then processed through two parallel embedding streams: a {global embedding} and a {per-series embedding} (see Fig.~\ref{figBD}). The global embedding, implemented via a shared linear layer, extracts dataset-wide features. In contrast, the per-series embedding layer is customized for each time series and captures localized information specific to the individual region. Both embeddings significantly reduce the dimensionality of the context vector, for example from 518 to 10. While the raw context RNN output is shared across all series, the embedding layers enable the primary RNN to receive a context vector that is dynamically adapted to the characteristics of the specific series being forecasted.

This form of series-specific contextualization is only feasible because the dataset is relatively small, allowing the use of separate embedding parameters per series. 
In earlier model variants, we attempted to reduce computational load by constructing the context batch from only a subset of “related” series, referred to as “friends” of the primary series. This approach is scalable to larger datasets, provided that relevant series can be efficiently identified. Two methods were explored: (i) selecting geographically adjacent regions, and (ii) using a data-driven selection based on Granger-causality-inspired modeling. In the latter, we evaluated pairwise predictive power between series and selected those that contributed most to reducing forecast error. Nevertheless, our experiments indicated that using the full batch with per-series embeddings consistently produced the best results.

\subsection{Primary and Context RNNs} \label{primaryAndContext}

\begin{figure}[t]
\centering
\includegraphics[width=0.7\textwidth]{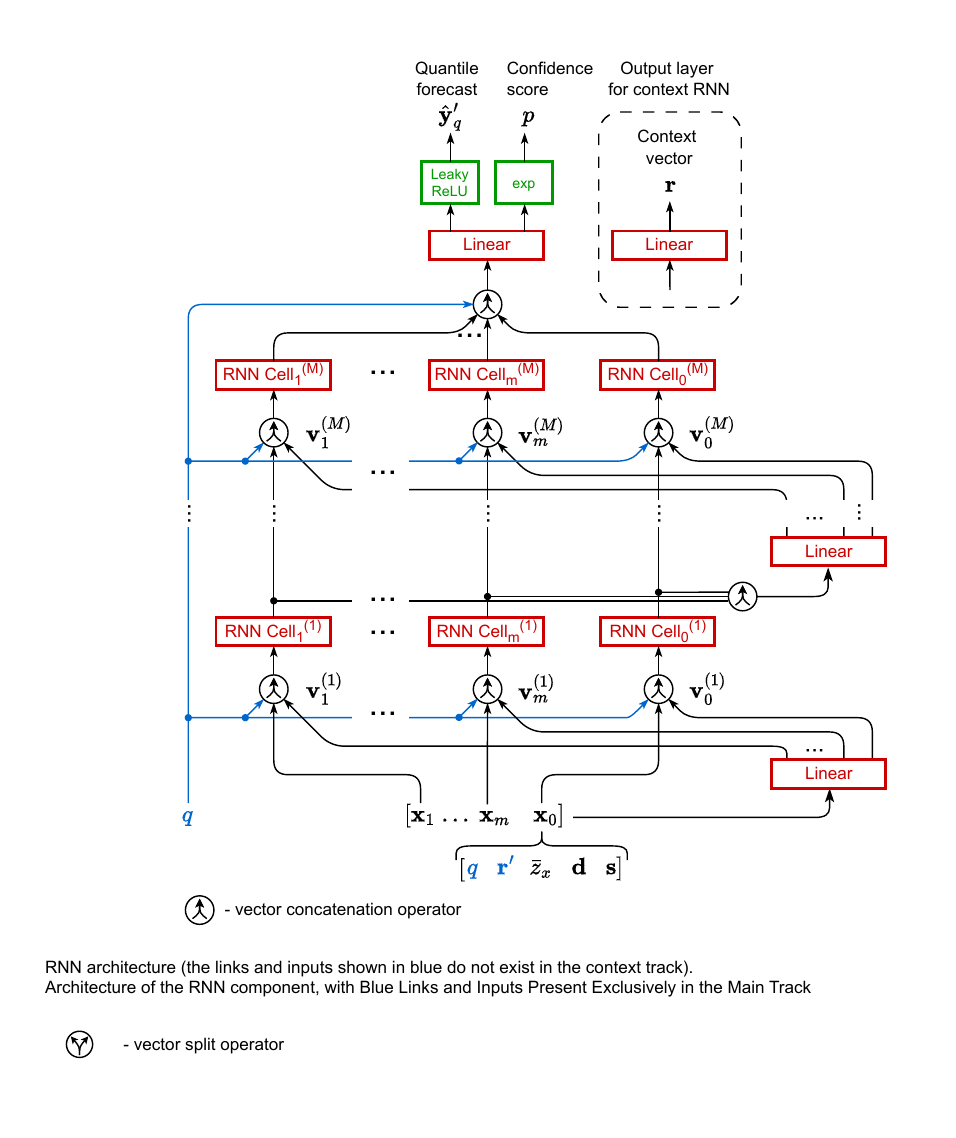}
\caption{Primary and context RNN architecture. Blue-highlighted inputs and connections are specific to the primary RNN and are absent in the context RNN.}
\label{figRNN}
\end{figure}

The architecture, illustrated in Fig.~\ref{figRNN}, is designed as a multi-layer RNN in which each layer comprises multiple dRNNCells and an associated context linear layer (this form of context should be distinguished from the context generated by the context track). A final linear layer aggregates the outputs of the last recurrent layer to generate either the model’s forecasts and confidence level (in the primary RNN) or the context vector (in the context RNN).

The solar power generation time series used in the experimental study is recorded at an hourly resolution. The RNN input primarily consists of sequences from the most recent $m$ days, i.e., $m \cdot 24$ hourly values (see Fig.~\ref{figSeg}). We found it beneficial to process each daily sequence (a patch of 24 values) semi-independently, using $m$ parallel streams, one for each day, while incorporating shared contextual information to preserve inter-day dependencies.

The model input includes $m$ daily patches, $\mathbf{x}_1, \dots, \mathbf{x}_m$, along with an additional patch $\mathbf{x}_0$, which aggregates auxiliary inputs. In the primary RNN, $\mathbf{x}_0$ includes the quantile level $q$, the context vector $\mathbf{r}'$, the mean value of the input segment $\bar{z}_x$, the date embedding $\mathbf{d}$, and a padding vector $\mathbf{s}$. 
In the context RNN, the quantile level $q$ and vector $\mathbf{r}'$ are omitted.

All components of $\mathbf{x}_0$ were described in the previous section, except for the padding vector $\mathbf{s}$, which adjusts the total length of $\mathbf{x}_0$ to match the dimensionality of the daily patches (i.e., 24). This padding ensures compatibility with the expected input structure of the recurrent cells, as explained in Section~\ref{IaO}. For example, if $q$ and $\bar{z}$ are scalars, $\mathbf{r}'$ is a 10-dimensional vector, and $\mathbf{d}$ has 3 dimensions, then $\mathbf{s}$ must be 9-dimensional. In the context RNN, where $q$ and $\mathbf{r}'$ are absent, $\mathbf{s}$ expands to 20 dimensions. The values in $\mathbf{s}$ are filled by randomly sampling from the remaining input components (indices are created during the network creation and used subsequently in all forward steps).

Each patch $\mathbf{x}_0, \dots, \mathbf{x}_m$ is augmented with a corresponding context vector $\mathbf{v}_0, \dots, \mathbf{v}_m$, which encodes a compressed representation of the entire input sequence. These context vectors are generated by a shared linear layer applied to the concatenated patches. Context layers are present in every recurrent layer, enabling the model to capture both local patterns within individual patches and global temporal dependencies across the full sequence.

The quantile level $q$ is also supplied to each recurrent layer as well as to the final output layer, ensuring that the forecast is explicitly conditioned on the requested quantile level.

The quantile forecasts produced by the primary RNN are passed through a Leaky ReLU activation function. This allows for near-zero outputs while attenuating negative values. During validation and testing, the forecasts are explicitly constrained to be non-negative, in order to maintain the physical plausibility of solar power generation values.

\subsection{Teams}

Each predictor is implemented as a team of RNNs (e.g., four models). All team members receive the same input, and each produces not only a $(h\cdot u)$-step forecast but also a scalar confidence score reflecting its certainty about the forecast. To generate the final team prediction, the forecasts of the top $K$ most confident members (e.g., $K = 3$) are selected, and their outputs are aggregated using the median.

Importantly, the selection of the top $K$ members is dynamic and may vary across different series and time steps. This design enables the emergence of "series- and situation-specific specialists," which can be interpreted as an automated and adaptive generalization of earlier approaches that partitioned datasets into subsets of similar time series, each modeled separately (see, e.g., \cite{Ban20}). 

The key advantage of this approach lies in its flexibility: the subdivision of series is not based on hand-crafted rules such as volatility or trend strength, nor does it rely on pre-defined clustering. Instead, the ML model autonomously learns grouping mechanisms in service of its primary objective -- forecast accuracy.

\subsection{Team Training}
\label{TT}

During training the quantile level $q$ is sampled from a Beta distribution ($\text{PDF} = q^{\alpha-1}(1-q)^{\beta-1}$) with both shape parameters $\alpha=\beta \in (0, 1)$, which results in oversampling small (near zero) and large (near one) quantile levels. This improves accuracy in these rapidly changing regions close to the boundary values of $q$. During testing, we sample $q$ from the grid $\Pi=\{0.001, 0.01, 0.02, ..., 0.99, 0.999\}$. For more details, see \cite{Smy26}.

Each team member produces not only a forecast but also a confidence score $p$, which is specific to the current time step and series. The model’s loss function is a weighted sum of two components: a standard quantile forecasting loss (pinball loss) and a confidence loss that regulates the reliability of the confidence scores. The overall loss is defined as:

\begin{equation}
L(\hat{y}'_{q},p) =
\begin{cases}
 L_{q}(y', \hat{y}'_{q}), & \text{if } \text{rank}(p)=\text{rank}(L_q), \\[6pt]
 L_{q}(y', \hat{y}'_{q}) + \gamma_1 p, & \text{if } \text{rank}(p)<\text{rank}(L_q) \quad \text{(overconfidence)}, \\[6pt]
 L_{q}(y', \hat{y}'_{q}) - \gamma_1 \gamma_2 p, & \text{if } \text{rank}(p)>\text{rank}(L_q) \quad \text{(underconfidence)}, \\
\end{cases}
\end{equation}
where $q \in (0,1)$ is the quantile level, $y' = y/\bar{z}_x$ is the normalized target, $\hat{y}'_{q}$ is the predicted normalized quantile, $\gamma_1$ and $\gamma_2$ are penalty coefficients (defined below), $\text{rank}(p)$ and $\text{rank}(L_q)$ are the confidence and accuracy loss ranks of the member (defined below) respectively, and $L_{q}(y', \hat{y}'_{q})$ denotes the pinball loss:

\begin{equation}
\label{eqrho}
L_q(y', \hat{y}'_q) = 
\begin{cases}
(y' - \hat{y}'_q)q, & \text{if } y' \geq \hat{y}'_q, \\[6pt]
(y' - \hat{y}'_q)(q - 1), & \text{if } y' < \hat{y}'_q.
\end{cases}
\end{equation}

The forward pass (i.e., computing outputs) involves all team members and is relatively efficient. In contrast, the backward pass (i.e., gradient computation) is computationally intensive. To improve efficiency and promote specialization, the backward pass is restricted -- using dynamic computation graphs (e.g., in PyTorch) -- to only the top $K$ most confident or most accurate team members at each time step. While this reduces computational cost, the primary motivation is to encourage the emergence of \textit{specialist} members that learn to recognize their own strengths and limitations, as reflected in their confidence scores.

The confidence loss for each member is determined by comparing its confidence rank, $\text{rank}(p)$, to its accuracy rank, $\text{rank}(L_q)$. To compute these ranks, the team members are sorted separately: in descending order by confidence scores and in ascending order by pinball losses. These two rankings should correlate well, and ideally, be the same. Overconfident members, whose accuracy rank exceeds their confidence rank, receive a positive penalty to reduce their confidence. Conversely, underconfident members are rewarded with a negative penalty, encouraging an increase in their confidence scores.

During training, the selection of the top $K$ members is based either on forecast accuracy or on confidence scores, with the choice between the two made randomly for each batch according to a predefined probability. During validation and testing, member selection is based exclusively on confidence scores.

Two additional automatic adjustments are applied during training:

\begin{enumerate}
    \item The ratio of the recent average forecasting loss to the confidence loss is regulated to remain close to a target value (e.g., $r=5$), thereby ensuring balance between the two objectives. To achieve this, after processing every $g$ (e.g., 100) batches the coefficient $\gamma_1$ is updated as:
    
    \begin{equation}
    \label{eqgamma}
        \gamma_1 = \frac{\bar{L_q}}{r \, \bar{p}^+}
    \end{equation}
    where $\bar{L_q}$ is the average pinball loss over the last $g$ batches, and $\bar{p}^+$ is the average confidence score for overconfidence cases ($\text{rank}(p) < \text{rank}(L_q)$) over the same period.

    \item The coefficient $\gamma_2$ is adjusted after processing every $g$ batches as follows:
    
    \begin{equation} \label{eq:lossesRatio}
        \gamma_2 = \gamma_2 + \operatorname{sign}(\bar{L_q})C
    \end{equation}
        where $C$ is constant, equal to 0.01.
              
\end{enumerate}

\subsection{Overlapping Quantile Subranges for Specialized Forecasting}
\label{ssect:overlapping}
The forecasting task involves predicting the next $h\cdot u$ hours of PV generation for an arbitrary quantile level $q \in (0,1)$. We observed that prediction accuracy improves when the full quantile range is partitioned into two or more subranges, with a separate predictor trained for each. Furthermore, performance can be enhanced by allowing these subranges to overlap and averaging the forecasts within the overlapping regions.

For example, consider two partitioning points (knots) at 0.2 and 0.6 with an overlap of 0.2, resulting in three overlapping subranges: $Q_1 = (0, 0.3)$, $Q_2 = (0.1, 0.7)$, and $Q_3 = (0.5, 1)$. A dedicated predictor is trained for each subrange. The final forecast for a given quantile $q$ is computed as follows:

\begin{equation}
\begin{split}
\text{for } q \in (0.0, 0.1]: & \quad \hat{\mathbf{y}}_q = \hat{\mathbf{y}}^1_q \\
\text{for } q \in (0.1, 0.3]: & \quad \hat{\mathbf{y}}_q = a \hat{\mathbf{y}}^1_q + (1 - a) \hat{\mathbf{y}}^2_q \\
\text{for } q \in (0.3, 0.5]: & \quad \hat{\mathbf{y}}_q = \hat{\mathbf{y}}^2_q \\
\text{for } q \in (0.5, 0.7]: & \quad \hat{\mathbf{y}}_q = b \hat{\mathbf{y}}^2_q + (1 - b) \hat{\mathbf{y}}^3_q \\
\text{for } q \in (0.7, 1.0): & \quad \hat{\mathbf{y}}_q = \hat{\mathbf{y}}^3_q
\end{split}
\end{equation}

Here, $\hat{\mathbf{y}}^1_q$, $\hat{\mathbf{y}}^2_q$, and $\hat{\mathbf{y}}^3_q$ denote the forecasts produced by the three predictors associated with the subranges $Q_1$, $Q_2$, and $Q_3$, respectively. The weights $a$ and $b$ are linear interpolation coefficients, computed as $\frac{q_u - q}{q_u - q_l}$, where $q_l$ and $q_u$ represent the lower and upper bounds of the overlapping region for a given pair of predictors. In our case:

\begin{equation}
a = -5q + 1.5, \qquad b = -5q + 3.5
\end{equation}

This structure allows the model to train "quantile-level specialists" -- predictors that are fine-tuned to perform well over specific segments of the quantile range. The ranges are intentionally asymmetric, as empirical results indicate that lower quantiles are more challenging to model accurately than upper quantiles.

\subsection{Implementation Details}

A naive implementation of our RNN architecture incurs significant computational overhead. Consider a configuration with three quantile-level ranges, four team members per range, and three patches: this setup yields 12 independent RNNs, requiring 36 cell computations per time step, multiplied by $M$ layers. To mitigate this inefficiency, we designed a higher-dimensional RNN architecture that consolidates these computations.

In the primary RNN, we extend standard LSTM-style cells from two-dimensional to four-dimensional parameter tensors. The modified dRNNCells are organized as follows:
\begin{itemize}
    \item first dimension: product of the number of quantile-level ranges and team size (e.g., $3 \cdot 4 = 12$),
    \item second dimension: number of patches (e.g., 3),
    \item additional dimensions: standard hidden state and cell state dimensions, as in LSTM cells.
\end{itemize}

Since the internal context mechanism operates at the patch level, the 4D RNN can be interpreted as a collection of 12 independent 3D RNNs. However, implementing them in a unified high-dimensional form yields substantial improvements in computational efficiency.

The context RNN employs a simplified 3D architecture with 3D dRNNCells, as it does not incorporate the team-based structure and does not require separate networks specialized for different quantile-level subranges.

\section{Empirical Results} \label{sec:results}

This section presents a comprehensive evaluation of the proposed AQ-model for short-term PV power forecasting across 259 European regions. The performance of the model is assessed in a probabilistic forecasting setting and compared against both statistical baselines and NN-based models.

\subsection{Data}

In this study, we employ the EMHIRES dataset for solar power generation \cite{Gon17}. The time series span 30 years (1986–2015), provide hourly resolution, and are available at several spatial aggregation levels defined by EUROSTAT. We use the NUTS-2 level, corresponding to basic administrative regions in Europe (typically provinces or other large sub-national units), and include data from 259 such regions.

The dataset contains hourly regional capacity factors, defined as the ratio of actual energy production to the maximum possible generation (installed capacity) in each region as of the end of 2015. These time series were derived by converting satellite-based solar radiation measurements using the Photovoltaic Geographic Information System (PVGIS) model. The generation process incorporates the geographical distribution of PV installations within each region, installed capacities, empirical peak-power and ramp characteristics, and duration curves. A comprehensive description of the methodology is available in \cite{Gon17}. The data were obtained from the EU JRC repository:
\url{https://data.jrc.ec.europa.eu/dataset/jrc-emhires-solar-generation-time-series}


\subsection{Performance Metrics}

The main performance metric used to evaluate quantile forecasts is the Continuous Ranked Probability Score (CRPS), defined as follows:

\begin{equation} \label{eqCRPS}
    {CRPS} = \frac{2}{|\Pi|}\sum_{q \in \Pi} L_{q}(y, \hat{y}_{q})
\end{equation}
where $\Pi$ is the set of quantile levels,  
$\hat{y}_{q}$ represents the predicted quantile at level $q$, and  
$L_{q}(y, \hat{y}_{q})$ is the pinball loss, as defined in~\eqref{eqrho}.

In addition, we use the Relative Frequency (RF) -- a calibration metric defined as follows: 
\begin{equation}
{RF}(q) = \frac{1}{N} \sum_{i=1}^N \mathds{1}{\{y_i \leq \hat{y}_{q,i}\}}
\label{eq:rf}
\end{equation}
where $N$ is the number of observations, $\hat{y}_{q,i}$ is the predicted $q$-quantile for the $i$-th observation, and $\mathds{1}$ denotes the indicator function.

The expected value of ${RF}(q)$ should equal the nominal probability level $q$. That is, the predicted $q$-quantile should exceed the true value in approximately $100q\%$ of cases, ensuring proper probabilistic calibration.  
To quantify deviations from this ideal behavior, we define the Relative Frequency Error (RFE) as:
\begin{equation}
{RFE}(q) = q - {RF}(q)
\label{eqRFE}
\end{equation}

Finally, the Mean Absolute RF Error (MARFE) aggregates deviations across all quantile levels $q \in \Pi$: 
\begin{equation}
{MARFE} = \frac{1}{|\Pi|} \sum_{q \in \Pi} |{RFE}(q)|
\label{eqMARFE}
\end{equation}

To evaluate the quality of prediction intervals (PIs), we apply the Winkler Score (WS), which balances two key aspects:  
sharpness -- how narrow the interval is, and calibration -- whether the observed value falls within the interval.  
The WS is defined as follows: 
\begin{equation}
{WS} =
\begin{cases}
(\hat{y}_{q_u} - \hat{y}_{q_l}) + \frac{2}{\alpha}(\hat{y}_{q_l} - y) & \text{if } y < \hat{y}_{q_l}, \\
(\hat{y}_{q_u} - \hat{y}_{q_l}) & \text{if } \hat{y}_{q_l} \leq y \leq \hat{y}_{q_u}, \\
(\hat{y}_{q_u} - \hat{y}_{q_l}) + \frac{2}{\alpha}(y - \hat{y}_{q_u}) & \text{if } y > \hat{y}_{q_u}
\end{cases}
\end{equation}
where $\hat{y}_{q_l}$ and $\hat{y}_{q_u}$ denote the predicted lower and upper quantiles defining the $100(1-\alpha)\%$ PI, with $\alpha = q_u - q_l$.

In this study, we evaluate 90\% PIs ($q_l = 0.05$ and $q_u = 0.95$) using the mean Winkler Score:
\begin{equation}
{MWS} = \frac{1}{N} \sum_{i=1}^{N} {WS}(y_i, \hat{y}_{q_l,i}, \hat{y}_{q_u,i})
\label{eqmws}
\end{equation}

To facilitate intuitive comparison of PI quality, we also calculate the percentages of observed values falling within, below, and above the 90\% PI. These metrics are compared against the target values -- in our case, 90\%, 5\%, and 5\%, respectively.

Finally, we assess the quality of point forecasts derived from the quantile forecasts. In this case, we assume that the point forecast is the predicted median $\hat{y}_{0.5}$. We define the Mean Absolute Error and Mean Squared Error based on the median as follows:

\begin{equation}
{MAE\text{-}Q} = \frac{1}{N} \sum_{i=1}^{N} | y_i - \hat{y}_{0.5,i}|
\label{eqmae}
\end{equation}

\begin{equation}
{MSE\text{-}Q} = \frac{1}{N} \sum_{i=1}^{N} (y_i - \hat{y}_{0.5,i})^2
\label{eqmse}
\end{equation}

{The selected evaluation metrics are directly linked to operational requirements in power system management. CRPS assesses the overall quality of probabilistic forecasts and is widely used to evaluate uncertainty-aware predictions. PI coverage and WS jointly reflect the trade-off between forecast sharpness and reliability, which is critical for reserve allocation and risk control. Finally, median-based error measures provide insight into the accuracy of point forecasts used for day-ahead scheduling and dispatch decisions.}

\subsection{Optimization, Training and Evaluation Setup} \label{SecTrain}

The data are partitioned chronologically. Model development is performed by training on all observations prior to 2010, validating and tuning hyperparameters on the 2010-2012 period, and subsequently retraining the model using the selected hyperparameters on data up to 2013. The final out-of-sample evaluation is performed on the 2013-2015 period (three years). The forecasting horizon is set to $h = 2$~days (48 hours).

Final predictions are produced using two variants: a single model (AQ-RNN) and an ensemble of five independently initialized models (AQ-RNNe). All ensemble members share the same architecture but differ in their random seeds and in the stochastic ordering of training samples. Ensemble forecasts are aggregated by taking the median across member predictions, resulting in robust and well-calibrated quantile estimates.

The AQ-RNN model is implemented in PyTorch and trained using the Adam optimizer. All AQ-RNN hyperparameters are selected by minimizing the CRPS on the validation set. Details of the hyperparameter selection are provided in \ref{app1}.

\subsection{Baseline Models} \label{SecBM}

The proposed AQ-RNN is evaluated against a set of reference probabilistic forecasting baselines, including both statistical and neural network-based methods. The selection and implementation of the baseline models follow the methodology described in our previous work~\cite{Smy26}. The considered baselines are summarized below:

\begin{itemize}
\item \textbf{ARIMA}, autoregressive integrated moving average model~\cite{Dud15}.
\item \textbf{Theta}, dynamic optimised Theta model, DOTM v3 from~\cite{Dud19}.
\item \textbf{MLP}, single-hidden-layer perceptron with sigmoid nonlinearities~\cite{Dud16}.
\item \textbf{DeepAR}, autoregressive RNN model for probabilistic forecasting~\cite{Sal20}.
\item \textbf{Transformer}, vanilla transformer~\cite{Vas17}.
\item \textbf{WaveNet}, deep autoregressive model combining causal filters with dilated convolutions~\cite{Oor16}.
\item \textbf{TFT}, temporal fusion transformer~\cite{Lim21}.
\end{itemize}

For the statistical models, ARIMA and Theta, the original hourly time series were decomposed into 24 separate sub-series, each corresponding to a specific hour of the day. Each sub-series was forecast two steps ahead, where each step represents a 24-hour interval, resulting in a final forecasting horizon of 48 hours.
The models generate point forecasts, with probabilistic outputs derived differently.
For ARIMA, we assume that the point forecast represents the mean of a normal distribution, and the standard deviation is estimated using the standard deviation of residuals \cite{Hyn21}.
Quantiles are obtained using R’s \texttt{forecast} package function \texttt{auto.arima()} with the \texttt{level} parameter set appropriately. The Theta method estimates quantiles through bootstrapping using the \texttt{dotm()} function in the \texttt{forecTheta} package~\cite{Fio16}.

Neural models used in this study come from GluonTS, a Python library for probabilistic time-series forecasting \cite{Ale20}. The MLP, DeepAR, and Transformer models generate forecasts parametrically: a projection layer outputs the parameters of a Student’s t-distribution (default), and model training minimizes its negative log-likelihood. Quantiles are then obtained directly from the learned parametric distribution.
WaveNET, in contrast, outputs a softmax-based categorical distribution over a fixed grid, modeling the autoregressive forecast probability directly.
TFT predicts multiple quantiles and a point forecast simultaneously; quantile outputs are produced by a linear layer on the decoder embedding. Training minimizes the aggregated quantile loss, and arbitrary quantiles are obtained via interpolation from the fixed grid.

\subsection{Results}

{
Table~\ref{tab1} compares the performance of the baseline models with the proposed AQ-RNN model and its ensemble variant, AQ-RNNe, which aggregates the forecasts of five individual AQ-RNN models. The performance metrics are reported on the test set for the quantile grid $\Pi=\{0.001, 0.01, 0.02, ..., 0.99, 0.999\}$ and are computed only for time steps with positive target values, excluding nighttime hours with zero PV generation.}

\begin{table}[]
  \caption{Summary of performance metrics (best values in bold, second-best values in italics)}
    \label{tab1}%
  \setlength{\tabcolsep}{3.4pt}
  \centering
    \begin{tabular}{lccccc}
	\toprule
    Model & \multicolumn{1}{c}{$CRPS$}        & \multicolumn{1}{c}{$MARFE$}      & \multicolumn{1}{c}{$MWS$}       & \multicolumn{1}{c}{$MAE$-$Q$}       & \multicolumn{1}{c}{$MSE$-$Q$}  \\
          & \multicolumn{1}{c}{$\times10^{-2}$}       & \multicolumn{1}{c}{$\times10^{-2}$}      & \multicolumn{1}{c}{$\times10^{-1}$}       & \multicolumn{1}{c}{$\times10^{-1}$}       & \multicolumn{1}{c}{$\times10^{-2}$}  \\
    \midrule
	ARIMA & 7.28$\pm$0.740 & 2.74$\pm$0.520 & 5.41$\pm$0.362 & 1.043$\pm$0.119 & 2.23$\pm$0.380 \\
    Theta & 7.34$\pm$0.735 & 2.53$\pm$0.475 & 5.49$\pm$0.372 & 1.044$\pm$0.120 & 2.30$\pm$0.387 \\
    DeepAR & 7.71$\pm$0.733 & 8.83$\pm$1.051 & 6.58$\pm$0.715 & 1.062$\pm$0.100 & 2.47$\pm$0.403 \\
    WaveNet & 7.57$\pm$0.654 & 4.81$\pm$2.536 & 6.33$\pm$0.441 & 1.060$\pm$0.108 & 2.56$\pm$0.431 \\
    Transformer & 9.71$\pm$0.869 & 19.62$\pm$2.407 & 9.59$\pm$0.992 & 1.288$\pm$0.113 & 3.58$\pm$0.460 \\
    TFT   & 6.95$\pm$0.632 & 2.99$\pm$1.283 & 4.91$\pm$0.344 & 1.007$\pm$0.106 & 2.12$\pm$0.373 \\
    AQ-RNN & \textit{6.17$\pm$0.724} & \textit{1.46$\pm$0.501} & \textit{3.96$\pm$0.301} & \textit{0.909$\pm$0.117} & \textit{1.87$\pm$0.371} \\
    AQ-RNNe & \textbf{6.09}$\pm$\textbf{0.728} & \textbf{1.21}$\pm$\textbf{0.544} & \textbf{3.91$\pm$0.293} & \textbf{0.896}$\pm$\textbf{0.117} & \textbf{1.81}$\pm$\textbf{0.364} \\
    \bottomrule
	\end{tabular}%
\end{table}%

\begin{figure}[t]
\centering
\includegraphics[width=0.4\textwidth]{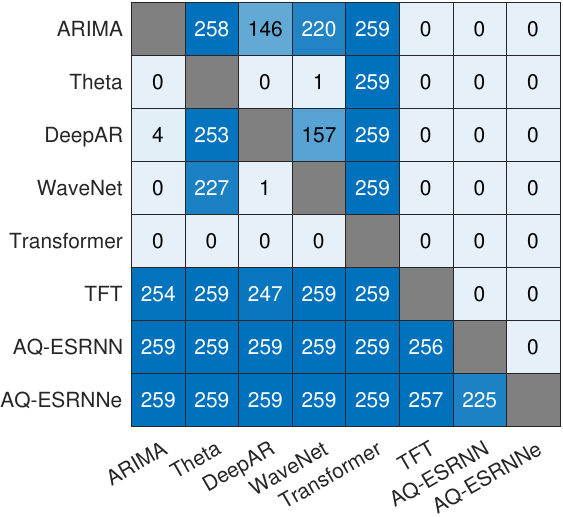}
\caption{Results of the Diebold-Mariano test: Each entry indicates the number of regions (out of 259) for which the model on the y-axis exhibits a significantly lower CRPS than the model on the x-axis, at the significance level $\alpha = 0.05$}\label{figDM}
\end{figure}

As shown in the table, AQ-RNN consistently yields lower prediction errors compared to the baselines. As expected, its ensemble version further improves the results.
To confirm these findings, a Diebold-Mariano test 
was conducted. This test assesses whether the CRPS value from two competing models differ significantly in expectation. The results, evaluated at a significance level of  $\alpha = 0.05$, are presented in Fig. \ref{figDM}. The test showed that AQ-RNN achieved statistically significantly lower CRPS values than all baseline models across all 259 regions, except when compared to TFT, which it outperformed in 256 regions. 
Notably, ensembling AQ-RNN further improved performance, yielding statistically significant gains over individual AQ-RNN models in 225 regions.

\begin{figure}[t]
\centering
\includegraphics[width=1\textwidth]{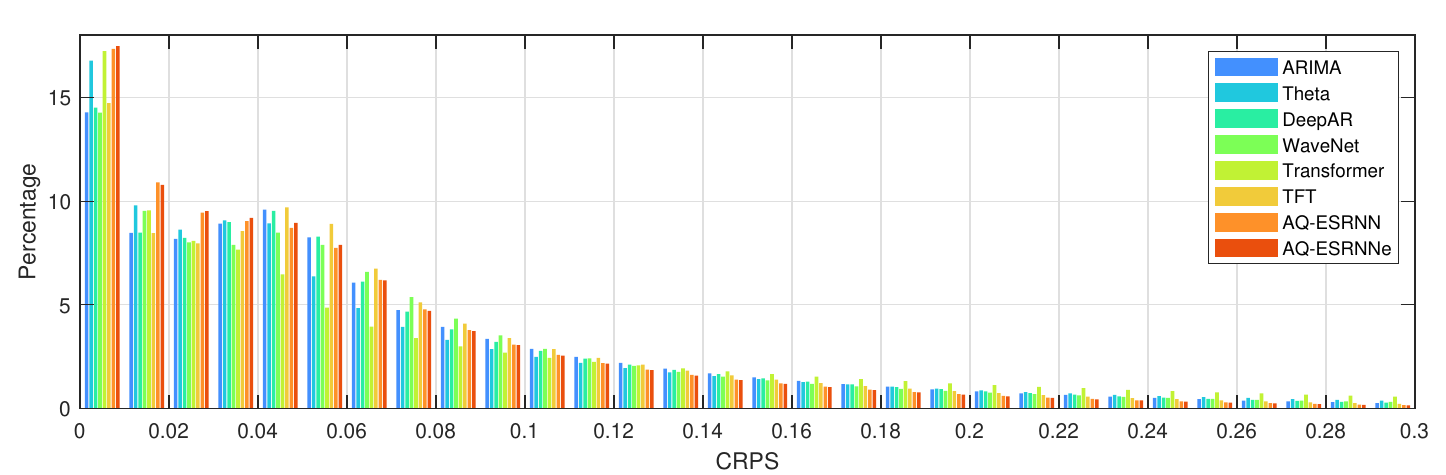}
\caption{CRPS value histograms across models.}\label{figHis}
\end{figure}

The distribution of CRPS values for all models is shown in the histogram in Fig.~\ref{figHis}. Note that AQ-RNN accounts for the highest percentage of the lowest CRPS values.

\begin{figure}[t]
\centering
\includegraphics[width=0.24\textwidth]{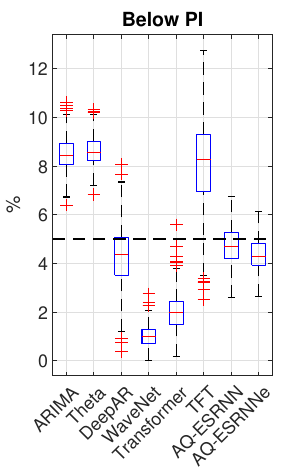}
\includegraphics[width=0.24\textwidth]{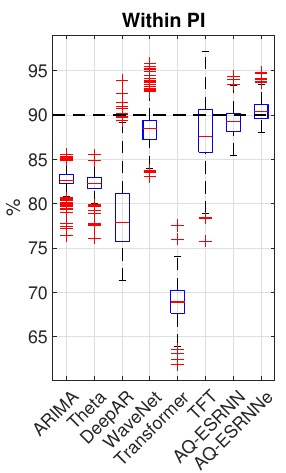}
\includegraphics[width=0.24\textwidth]{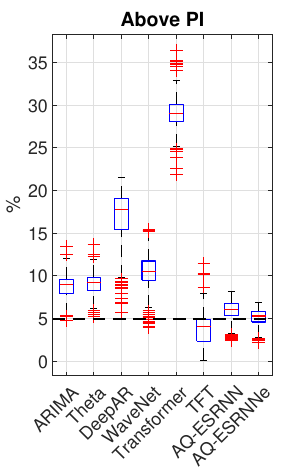}
\caption{Percentage of forecasts below, within and above the 90\% PI. The dashed line represents the target value.}\label{fig2}
\end{figure}

Fig. \ref{fig2} shows the percentage of forecasts falling below, within, and above the 90\% PI. Notably, our model produces results that are very close to the target coverage levels, outperforming all other models in this respect.

\begin{figure}[t]
\centering
\includegraphics[height=0.25\textwidth]{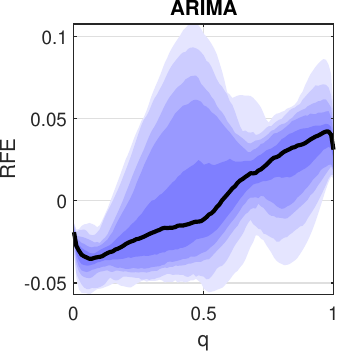}
\includegraphics[height=0.25\textwidth]{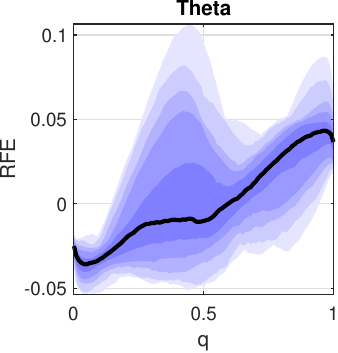}
\includegraphics[height=0.25\textwidth]{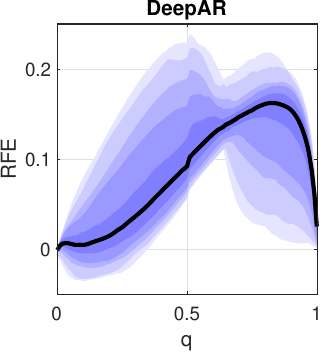}
\includegraphics[height=0.25\textwidth]{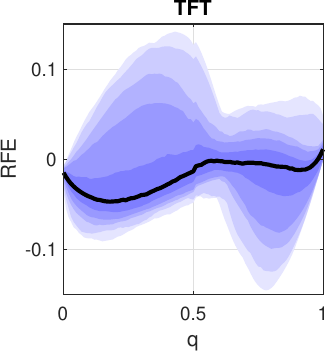}
\includegraphics[height=0.25\textwidth]{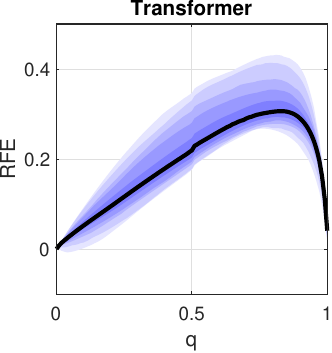}
\includegraphics[height=0.25\textwidth]{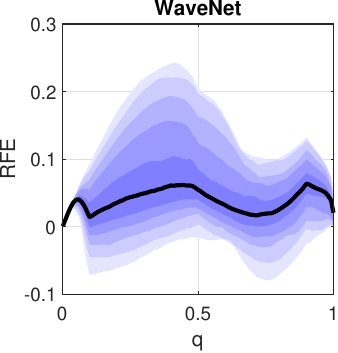}
\includegraphics[height=0.25\textwidth]{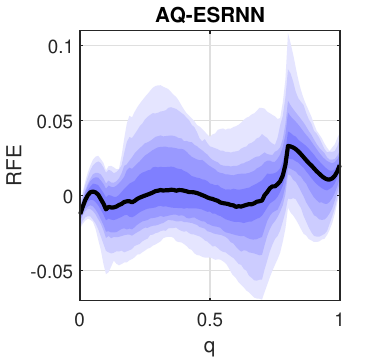}
\includegraphics[height=0.25\textwidth]{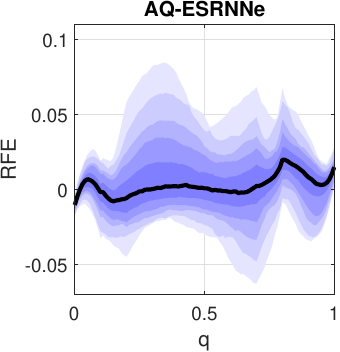}
\caption{Relative Frequency Error (quantiles 0.001, 0.01, 0.05, 0.1, 0.25, 0.5 (black line), 0.75, 0.9, 0.95, 0.99, and 0.999 are shown)}\label{figRFE}
\end{figure}

RFE across quantile levels $q \in \Pi$ is shown in Fig. \ref{figRFE}. From this figure, we observe that ARIMA and Theta exhibit very similar RFE distributions, which result from their comparable approach to calculating PV generation quantiles based on residuals (see Section\ref{SecBM}). For both models, the predicted quantile levels tend to be overestimated for lower $q$ values and underestimated for higher ones. In the case of DeepAR, Transformer, and WaveNet, the predicted quantiles are generally underestimated, with the largest RFE observed for the Transformer model. TFT, on the other hand, tends to slightly overestimate the predicted quantile levels. AQ-RNN demonstrates the most accurate performance -- the predicted quantiles are closest to the expected values, and the RFE distribution is the narrowest compared to the baseline models. Its ensemble variant, AQ-RNNe, further improves the results, particularly around $q = 0.8$.

\begin{figure}[t]
\centering
\includegraphics[width=0.24\textwidth]{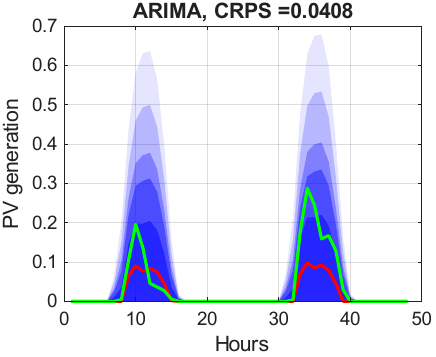}
\includegraphics[width=0.24\textwidth]{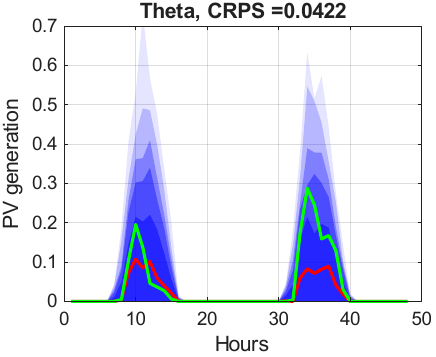}
\includegraphics[width=0.24\textwidth]{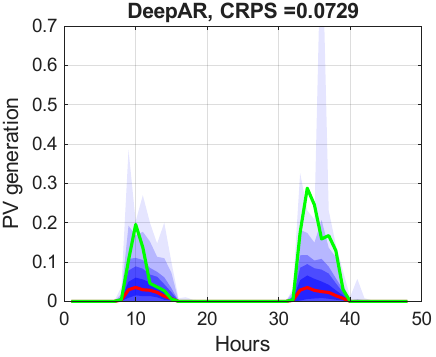}
\includegraphics[width=0.24\textwidth]{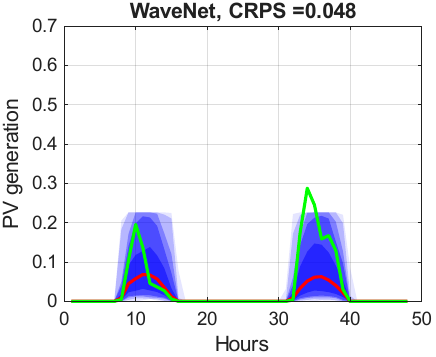}
\includegraphics[width=0.24\textwidth]{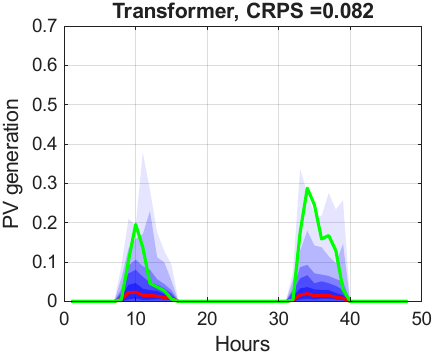}
\includegraphics[width=0.24\textwidth]{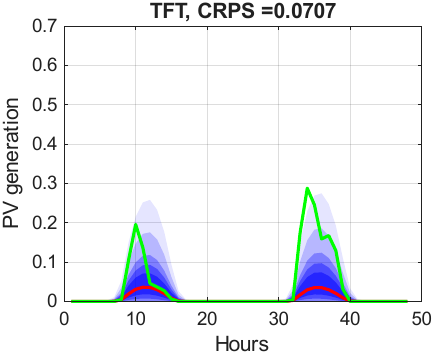}
\includegraphics[width=0.24\textwidth]{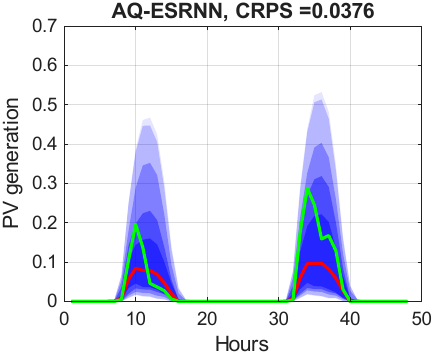}
\includegraphics[width=0.24\textwidth]{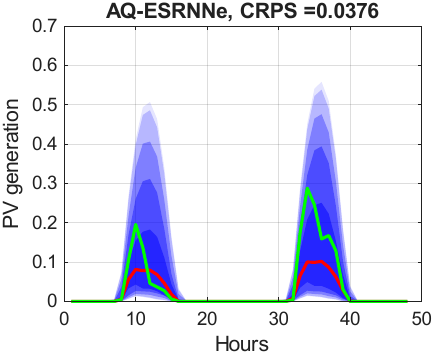}
\includegraphics[width=0.9\textwidth]{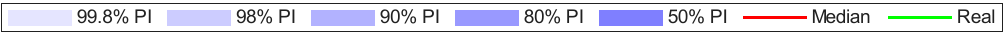}
\caption{Examples of quantile forecasts for region PL11, 1 and 2 January 2015.}
\label{figF1}
\end{figure}

\begin{figure}[t]
\centering
\includegraphics[width=0.24\textwidth]{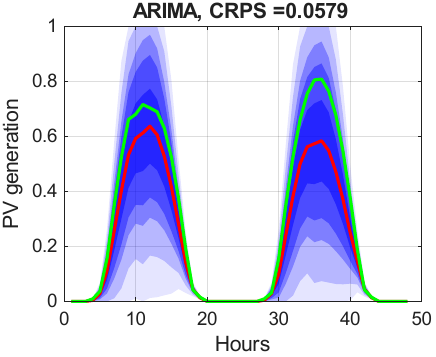}
\includegraphics[width=0.24\textwidth]{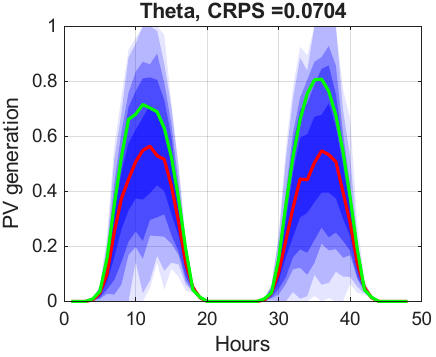}
\includegraphics[width=0.24\textwidth]{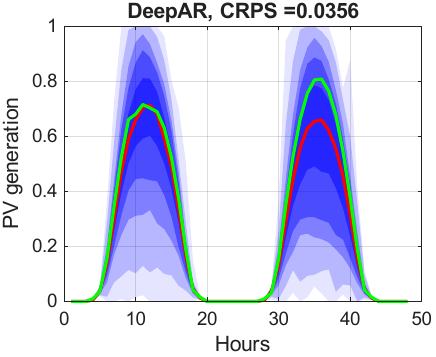}
\includegraphics[width=0.24\textwidth]{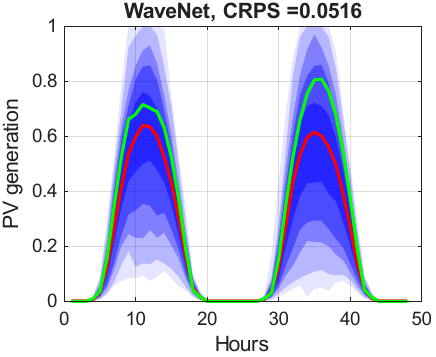}
\includegraphics[width=0.24\textwidth]{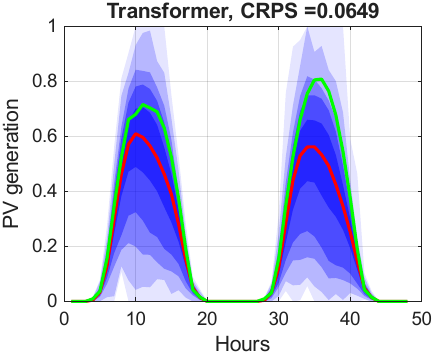}
\includegraphics[width=0.24\textwidth]{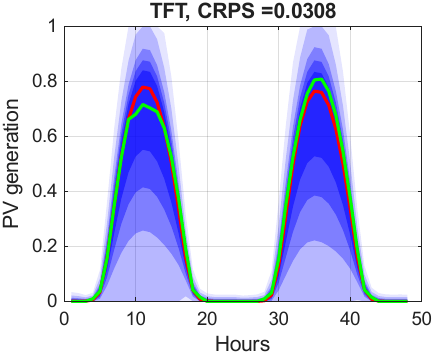}
\includegraphics[width=0.24\textwidth]{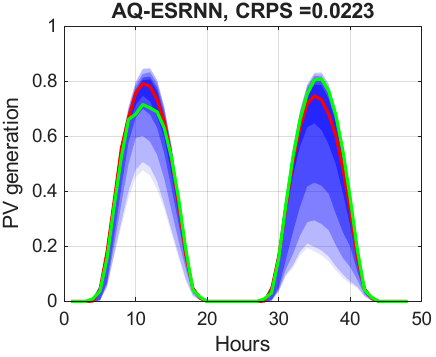}
\includegraphics[width=0.24\textwidth]{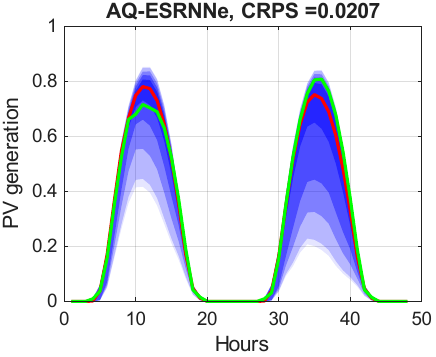}
\includegraphics[width=0.9\textwidth]{figs/Legend.pdf}
\caption{Examples of quantile forecasts for region PL11, 1 and 2 July 2015.}
\label{figF2}
\end{figure}

Examples of PV generation quantile forecasts for region PL11, covering winter (1-2 January 2015) and summer (1-2 July 2015), are shown in Figs.~\ref{figF1} and \ref{figF2}, respectively. The CRPS metric indicates that AQ-RNN provides the most accurate probabilistic forecasts for these cases.

\subsection{Ablation Studies}
\label{ssec:ablations}
Ablation study results are presented in Table~\ref{tab_ab}, with the full model shown in the last row. We evaluate the following reduced variants of AQ-RNNe:

\begin{itemize}
\item {No global context adapter / No per-series adapter:} Removing the global context adapter has a larger negative impact on accuracy than removing the per-series adapters (see Fig.~\ref{figBD}).

\item {Global$\rightarrow$per-series adapter:} Here, per-series adapters operate on the output of the global adapter. Although seemingly reasonable, this modification decreases accuracy.

\item {No context:} Complete removal of all context leads to the most severe accuracy drop among the above variants, as expected.

\item {No patches:} Patching has a positive effect, although the improvement is modest.

\item {No teams:} The team mechanism appears to be one of the most influential components.

\item {No quantile subranges:} Removing the overlapping quantile subranges is neutral or even slightly beneficial on average; however, performance deteriorates at the distribution edges (lower and upper quantiles), as shown in Table~\ref{tab_edge_full}.
\end{itemize}

\begin{table}[]
  \caption{Ablation study results (5 members ensemble)}
    \label{tab_ab}
  \setlength{\tabcolsep}{3.4pt}
  \centering
    \begin{tabular}{lccccc}
	\toprule
    Model & \multicolumn{1}{c}{$CRPS$}        & \multicolumn{1}{c}{$MARFE$}      & \multicolumn{1}{c}{$MWS$}       & \multicolumn{1}{c}{$MAE$-$Q$}       & \multicolumn{1}{c}{$MSE$-$Q$}  \\
          & \multicolumn{1}{c}{$\times10^{-2}$}       & \multicolumn{1}{c}{$\times10^{-2}$}      & \multicolumn{1}{c}{$\times10^{-1}$}       & \multicolumn{1}{c}{$\times10^{-1}$}       & \multicolumn{1}{c}{$\times10^{-2}$}  \\
    \midrule
	No global cntx adapter & 6.47 & 1.29 & 4.04 & 0.961 & 2.04 \\
    No per-series cntx adapter & 6.31 & 1.57 & 4.03 & 0.933 & 1.94 \\
    Global$\rightarrow$per-series adapter & 6.15 & 1.42 & 3.96 & 0.908 & 1.86 \\
    No context & 6.55 & 1.63 & 4.10 & 0.975 & 2.11 \\
    No patches & 6.11 & 1.36 & 3.92 & 0.903 & 1.82 \\
    No teams & 9.03 & 9.13 & 8.43 & 1.240 & 2.58 \\
    No quantile subranges & 6.08 & 1.16 & 3.93 & 0.895 & 1.81 \\
    AQ-RNNe & {6.09}& {1.21} & {3.91}& {0.896}& {1.81} \\
    \bottomrule
	\end{tabular}%
\end{table}%

\begin{table}[]
  \setlength{\tabcolsep}{7pt}
  \caption{Performance metrics of the full model (left) versus the model without quantile subranges (right) evaluated at the boundary quantile levels}
    \label{tab_edge_full}
  \centering
    \begin{tabular}{lccccc}
	\toprule
     \multicolumn{1}{c}{Quantile level}        & \multicolumn{1}{c}{Mean quantile loss}      & \multicolumn{1}{c}{Mean RFE Bias}   \\
     \multicolumn{1}{c}{}       & \multicolumn{1}{c}{$\times10^{-3}$}      & \multicolumn{1}{c}{$\times10^{-1}$}    \\
    \midrule
	0.001 & $0.41 / 0.56$ & $-1.01/-1.82$ \\
    0.01 & 2.32/2.43 & $-0.49/-1.29$ \\
    0.02 & 4.35/4.43 & $0.003/0.75$ \\
    .. & .. & .. \\
    0.98 & 4.20/4.22 & 0.63/0.85 \\
    0.99 & 2.27/2.30 & 1.03/1.36 \\
    0.999 & 0.46/0.52 & 1.51/1.88 \\
    \bottomrule
	\end{tabular}%
\end{table}%

\section{Discussion}
{
This study investigates probabilistic PV power forecasting from a system-level perspective, with particular emphasis on uncertainty quantification, spatial aggregation, and operational relevance. The empirical results demonstrate that the proposed AQ-RNN framework delivers consistent improvements over established statistical and neural baselines in a large-scale, multi-regional setting. Beyond overall accuracy gains, several methodological and operational aspects of the proposed approach merit further discussion in the context of power system operation and energy management.

\subsection{Methodological considerations}

Several methodological aspects of the proposed framework warrant additional discussion. During both training and inference, the model operates on the full 24-hour daily cycle. However, when computing performance metrics, time steps with zero actual generation are excluded, as they correspond to nighttime conditions and are perfectly predictable. The final non-linearity of the model, implemented as a Leaky ReLU followed by truncation of negative forecasts to zero, ensures physically meaningful outputs and stable behavior in these regimes. Nevertheless, the potential impact of explicitly modifying the learning objective to account for zero-generation periods -- such as by zeroing the loss during nighttime hours or training on a positive-only dataset -- has not been explored and represents an interesting direction for future research.

The context adaptation mechanism also raises important considerations regarding parameterization. The per-series context adapters consist of relatively large parameter matrices; however, empirical results indicate that training remains stable and does not lead to overfitting. At the same time, the ablation analysis shows that the global context adapter is not redundant and, in fact, plays a more critical role than series-specific adapters. This suggests that dataset-wide contextual information is a primary driver of performance gains, while per-series adaptations provide complementary refinements.

The team-based ensemble mechanism emerges as one of the most influential components of the proposed architecture. Its removal leads to the largest deterioration across all accuracy and uncertainty metrics. Interestingly, the mechanism performs optimally only for a specific configuration (three selected members out of four), while alternative configurations (e.g., 2/3 or 3/5) result in substantially worse performance. Although this behavior is consistent across experiments, its underlying causes are not yet fully understood. Further investigation is needed to clarify the interaction between ensemble size, specialization, and confidence estimation, and to determine whether adaptive or theoretically grounded team selection strategies could further improve robustness.

Finally, although splitting the quantile range into overlapping subranges improves calibration at very low and very high probability levels, performance at the distributional extremes remains challenging. One possible explanation is the relatively small magnitude of the pinball loss near the boundaries, particularly for lower quantiles where both forecasts and observations are constrained to be non-negative. Future work could explore loss reweighting schemes or alternative representations of the quantile level to further enhance calibration in these regions. In addition, while the current aggregation strategy relies on the median of the selected team members for robustness, confidence-weighted averaging may offer further improvements and warrants investigation.

\subsection{Benefits of any-quantile probabilistic forecasting}

A key advantage of the proposed framework lies in its ability to generate calibrated forecasts at arbitrary quantile levels within a single trained model. Unlike conventional probabilistic approaches constrained to fixed quantile grids or parametric distributional assumptions, the any-quantile formulation provides enhanced flexibility for operational decision-making. System operators can directly extract quantile levels tailored to specific risk preferences, reserve requirements, or market mechanisms without retraining or post-processing. This capability is particularly relevant in power systems with high renewable penetration, where uncertainty characteristics vary dynamically across time, regions, and operating conditions.

The empirical results indicate that this flexibility does not come at the expense of forecast quality. On the contrary, AQ-RNN achieves improved calibration and sharper prediction intervals, as reflected by lower CRPS and Winkler scores. These improvements translate into more reliable uncertainty estimates, which are critical for balancing the trade-off between operational security and economic efficiency.

\subsection{Role of cross-regional context in uncertainty reduction}

The ablation analysis highlights the importance of cross-regional contextual information for probabilistic PV forecasting. Removing the context mechanism leads to a clear deterioration in calibration and interval quality, indicating that spatial dependencies play a significant role in reducing forecast uncertainty at the system level. This finding is consistent with the physical characteristics of solar generation, where weather patterns and cloud movements exhibit spatial coherence across neighboring and even distant regions.

By explicitly modeling these inter-regional dependencies, the proposed framework effectively leverages information from geographically distributed PV assets, improving robustness in both high-variability and low-generation regimes. From an operational standpoint, this suggests that portfolio-level forecasting approaches that exploit spatial structure are better suited for regional energy management than isolated single-site models.

\subsection{Importance of team-based ensembles for robustness}

Another important insight from the ablation study concerns the role of the team-based ensemble mechanism. Disabling this component results in the largest degradation across all error metrics, underscoring its critical contribution to forecasting performance. The dynamic selection and aggregation of specialized predictors enhances robustness against diverse operating conditions, including seasonal effects, regime shifts, and extreme events.

Rather than relying on a single monolithic predictor, the team-based structure enables adaptive specialization, allowing different model instances to contribute where they are most confident. This mechanism improves not only average accuracy but also stability and reliability, which are essential properties for operational deployment in power systems subject to non-stationary behavior.

\subsection{Implications for power system operation and future research}

From an energy management perspective, the improvements achieved by AQ-RNN have direct practical implications. Better-calibrated probabilistic forecasts support more accurate reserve sizing, reduce the likelihood of costly over- or under-provisioning, and enhance short-term scheduling decisions. In systems with limited flexibility and reduced inertia, such improvements contribute to maintaining reliability while minimizing operational costs.

Importantly, the proposed framework operates solely on historical generation data, avoiding dependence on high-resolution numerical weather predictions that may be unavailable, inconsistent, or unreliable across long time horizons and large spatial domains. This characteristic increases the robustness and transferability of the approach, making it applicable in a wide range of real-world settings.

While the results are encouraging, several limitations warrant discussion. The current study focuses on PV generation-only inputs and does not incorporate exogenous meteorological variables. Although this choice enables a consistent large-scale evaluation, future work could investigate different configurations that integrate weather forecasts where reliable data are available. In addition, extending the framework to finer spatial resolutions, mixed renewable portfolios, or explicitly quantifying economic benefits within market-clearing or unit commitment frameworks represents promising directions for further research.}

\section{Conclusion}

{This paper introduced an any-quantile probabilistic forecasting framework for multi-regional photovoltaic power generation based on the proposed Any-Quantile Recurrent Neural Network (AQ-RNN). 
The approach addresses the limitations of deterministic forecasting by providing flexible and calibrated uncertainty estimates that better support operational decision-making in power systems with high renewable penetration.

The proposed model 
integrates cross-regional context information, dynamic team-based learning, and efficient quantile-level specialization, thereby enhancing both statistical accuracy and practical utility. It
enables the estimation of conditional quantiles at arbitrary probability levels within a single trained network and exploits cross-regional information to improve system-level robustness. Large-scale empirical evaluation across 259 European regions demonstrates consistent improvements over established probabilistic benchmarks in terms of forecast accuracy and calibration, highlighting the practical relevance of the approach for uncertainty-aware energy management.
}

--------------------------------------------------------------------------------------





\appendix
\section{Hyperparameter Configuration of AQ-RNN}
\label{app1}

The hyperparameters are set as follows:

\begin{itemize}
    \item \texttt{Batch size} -- an increasing batch size schedule is used. Training starts with a batch size of 2, which is increased to 5 in epoch~2, to 12 in epoch~3, and finally to 25 in epoch~4. This schedule is denoted as 0:2, 2:5, 3:12, 4:25 in Table~\ref{table:esrnn_hyperparameter_settings}.

    \item \texttt{Updates per epoch} -- the number of updates per epoch decreases as the batch size increases, starting from 8320 updates and ending at approximately 3920 updates per epoch for a batch size of 25. A sublinear formula is employed to keep the execution time per epoch approximately constant; for details, see~\cite{Smy26}.

    \item \texttt{Epochs} -- eight epochs are used. Increasing the number of epochs to nine occasionally yields minor improvements, but the gains do not justify the additional computational cost.

    \item \texttt{Learning rate} -- in addition to batch size scheduling, learning rate scheduling is applied. Training starts with a learning rate of 0.001, which is reduced by a factor of 3 in epoch~5, then to $1/8$ of the original rate in epoch~6, and finally to $1/20$ of the original rate in epoch~7. This schedule is denoted as 0:0.001, 5:/3, 6:/8, 7:/20 in Table~\ref{table:esrnn_hyperparameter_settings}.


    \item \texttt{LR multiplier} -- a scaling factor applied to the learning rate of per-series parameters, here the context adapters (embeddings); see Section~\ref{context}. Higher learning rates are used for per-series parameters to compensate for their less frequent updates, as these parameters are adjusted only when a given series appears in the training batch.

    \item \texttt{Beta shape} ($\alpha=\beta$) -- the shape parameter of a symmetric Beta distribution used to sample quantile levels during training, see Section \ref{TT}.

    \item \texttt{Dilations} -- define the AQ-RNN architecture. Each inner square bracket lists the dilations of layers within a given block. For example, \texttt{[[2], [4], [8]]} denotes a model composed of three blocks, each with a single layer ($M=3$) and progressively increasing delays of 2, 4, and 8 daily time steps. Two-layer variants (e.g., \texttt{[[2], [4]]}) and four-layer variants (e.g., \texttt{[[2], [4], [8], [3]]}, \texttt{[[2], [4], [8], [16]]}) were also evaluated. The latter produced results comparable to \texttt{[[2], [4], [8]]}, but with higher computational cost.

    \item \texttt{Training steps} -- the number of recurrent steps over which the RNN is unrolled after initialization from a randomly sampled starting point for each series and after executing warmup steps; see~\cite{Smy24a}.

    \item \texttt{Criterion selection probability} -- the probability that determines one of two ways of selecting team members during training, based either on the accuracy or on the confidence; see Section~\ref{TT}.

    \item \texttt{Desired losses ratio} ($r$) -- the target average ratio between accuracy-related losses and confidence-related losses; see \eqref{eqgamma}.


    \item \texttt{Per-series context length} -- the length of the vector $\mathbf{r}'$, corresponding to the output dimensionality of the context RNN for each context batch member (with $L = 259$ series); see Section~\ref{context}.

    \item \texttt{Embedded context length} -- the output dimensionality of the per-series embedding layer applied to the flattened vector of size $L \cdot$\texttt{Per-series context length}; see Section~\ref{context}.


    \item \texttt{Input window length} ($m$) -- the number of days used to construct the input vector $\textbf{x}$; see Fig.~\ref{figSeg}.

    \item \texttt{Patch context size} and \texttt{Patch h-state size} -- parameters related to the daily patch RNNs; see Section~\ref{primaryAndContext}. The former specifies the dimensionality of each $\mathbf{v}_i$ vector, while the latter defines the size of the corresponding hidden state, as illustrated in Fig.~\ref{figRNN}. The size of the c-state equals the sum of the \texttt{Patch h-state size} and the \texttt{Cell output size}, which is set to 24.

    \item \texttt{Team configuration} -- it is encoded by specifying number of top members followed by number of all members, e.g. 3/4 denotes 4-member team with 3 top members.


    \item \texttt{Date embedding size} -- the dimensionality of the vector $\mathbf{d}$; see Section~\ref{IaO}. The forecast origin (i.e., the last date in the input window) is first encoded as a one-hot vector of length 52 representing the week of the year (with the few days of week 53 mapped to week 52). This representation is then embedded into the vector $\mathbf{d}$. The day-of-week information is not explicitly modeled, as the focus is on natural rather than human-driven processes.


    \item \texttt{Update frequency} ($g$) -- the update frequency (in batches) for the automatic adjustment of $\gamma_1$ and $\gamma_2$ according to \ref{eqgamma} and \ref{eq:lossesRatio}.


    \item \texttt{Quantile ranges} -- as explained in Section~\ref{ssect:overlapping}, the unit interval of quantile levels is divided into overlapping subranges. For example, three ranges defined by two knots at 0.2 and 0.6 with an overlap of 0.1 are encoded as $0.2,\,0.6 \pm 0.1$.

\end{itemize}

\begin{table*}[!t]
    \centering
    \caption{Settings of AQ-RNN hyperparameters and the hyperparameter search grid. See Section~\ref{SecTrain}}
    \label{table:esrnn_hyperparameter_settings}

    \begin{tabular}{lcc}
        \toprule
        Hyperparameter & Value  & Grid \\ 
        \midrule
        \texttt{Batch size} & 0:2, 2:5, 3:12, 4:25,  & also 0:2, 2:5, 3:12\\ 
        \texttt{Updates per epoch} & 8320, ..., 3920 & * \\
        \texttt{Epochs}	& 8 & 4, ..., 9	\\ 
        \texttt{Learning rates} & 0:0.001, 5:/3, 6:/8, 7:/20 & also constant 0.003 \\        
        \texttt{LR multiplier} & 3 & 1,3,10  \\
        \texttt{Beta shape} & 0.5 & 0.3,0.5,0.7	\\
        \texttt{Dilations} & [[2],[4],[8]] & *  \\
        \texttt{Training steps} & 20 & 10, 20, 30, 50  \\
        \texttt{Crit. sel. probability} & 0.9 & 0.5, 0.65, 0.9,\\
        && 0.95, 0.99  \\
        \texttt{Desired losses ratio} & 5 & 1,3,5,10  \\
        \texttt{Per-series context length} & 2 & 1,2,4 \\
        \texttt{Embedded context length} & 10 & 5,10,20 \\
        \texttt{Input window length} & 4 & 3,4,5 \\
        \texttt{Patch context size} & 5 & 2,5,10 \\
        \texttt{Patch h-state size} & 5 & 5,10 \\
        \texttt{Team configuration*} & 3/4 & 1/2, 1/3, 2/3, 2/4,  \\
        && 3/4, 2/5, 3/5, 4/5\\
        \texttt{Date embed. size} & 3 & 3,5 \\
        \texttt{Update frequency} & 20 & 20,50 \\
        \texttt{Quantile ranges} & $0.2,0.6 \pm 0.1$ & several, including \\
        &&$0.3,0.7 \pm 0.1$ \\

        \bottomrule
        * -- see description in the text
    \end{tabular}
\end{table*}







\bibliographystyle{plain}

\bibliography{main}




\section*{Declaration of generative AI and AI-assisted technologies in the manuscript preparation process}

During the preparation of this work the authors used ChatGPT, Claude and Gemini in order to improve the language and clarity of the text. After using this tool/service, the authors reviewed and edited the content as needed and take full responsibility for the content of the published article.

\end{document}